\documentclass{article}

\usepackage{microtype}
\usepackage{graphicx}
\usepackage{subfigure}
\usepackage{booktabs} 

\usepackage{hyperref}

\usepackage[accepted]{icml2024}

\usepackage{amsmath}
\usepackage{amssymb}
\usepackage{mathtools}
\usepackage{amsthm}

\usepackage[capitalize,noabbrev]{cleveref}

\theoremstyle{plain}

\theoremstyle{definition}

\theoremstyle{remark}

\usepackage[textsize=tiny]{todonotes}

\icmltitlerunning{Fine-tuned network relies on generic representation to solve unseen cognitive task}

\begin{document}

\twocolumn[
\icmltitle{Fine-tuned network relies on generic representation to solve unseen cognitive task}




\begin{icmlauthorlist}
\icmlauthor{Dongyan Lin}{x,y}
\end{icmlauthorlist}

\icmlaffiliation{x}{Mila}
\icmlaffiliation{y}{McGill University, Montreal, Canada}

\icmlcorrespondingauthor{Dongyan Lin}{dongyan.lin@mail.mcgill.ca}

\icmlkeywords{Transformers, fine-tuning, computational neuroscience, cognitive science}

\vskip 0.3in
]



\printAffiliationsAndNotice{}  
\setlength{\parskip}{0.1em}
\begin{abstract}
Fine-tuning pretrained language models has shown promising results on a wide range of tasks, but when encountering a novel task, do they rely more on generic pretrained representation, or develop brand new task-specific solutions? Here, we fine-tuned GPT-2 on a context-dependent decision-making task, novel to the model but adapted from neuroscience literature. We compared its performance and internal mechanisms to a version of GPT-2 trained from scratch on the same task. Our results show that fine-tuned models depend heavily on pretrained representations, particularly in later layers, while models trained from scratch develop different, more task-specific mechanisms. These findings highlight the advantages and limitations of pretraining for task generalization and underscore the need for further investigation into the mechanisms underpinning task-specific fine-tuning in LLMs. 
\end{abstract}

\section{Introduction}
\label{introduction}

Pretraining large language models (LLMs) on vast amounts of data has shown promising results, revealing emerging capabilities such as mathematical skills, translation, and reasoning. This extensive pretraining helps models develop useful representations that allow for subsequent fine-tuning on specific, unseen tasks, thereby broadening the range of problems LLMs can solve. However, it remains unclear whether a pretrained model, when addressing specific tasks, relies on generic pretrained representations that are broadly applicable to most language-based tasks or if it constructs new, niche representations uniquely suited to the target task, akin to optimizing the network from scratch for that task.

In this work, we investigate the ability of pretrained GPT-2 to solve a context-dependent decision-making problem based on numerical comparison through fine-tuning. This task is adapted from neuroscience and cognitive science literature, commonly used to study decision-making in animals and recurrent neural network (RNN) models of the brain, but it is entirely novel to GPT models as it is not present in English text datasets. We aim to understand the extent to which fine-tuned models depend on their pretrained representations to solve a novel task. To this end, we compare the representations after fine-tuning with those developed by GPT-2 optimized solely on this task from scratch. We chose this task not only because it is novel but also because its grounding in neuroscience allows us to explore the data with computational neuroscience methods and make direct comparisons between representations in biological and artificial neural networks.
 \begin{figure}[h]
    \centering
     \includegraphics[scale=0.63]{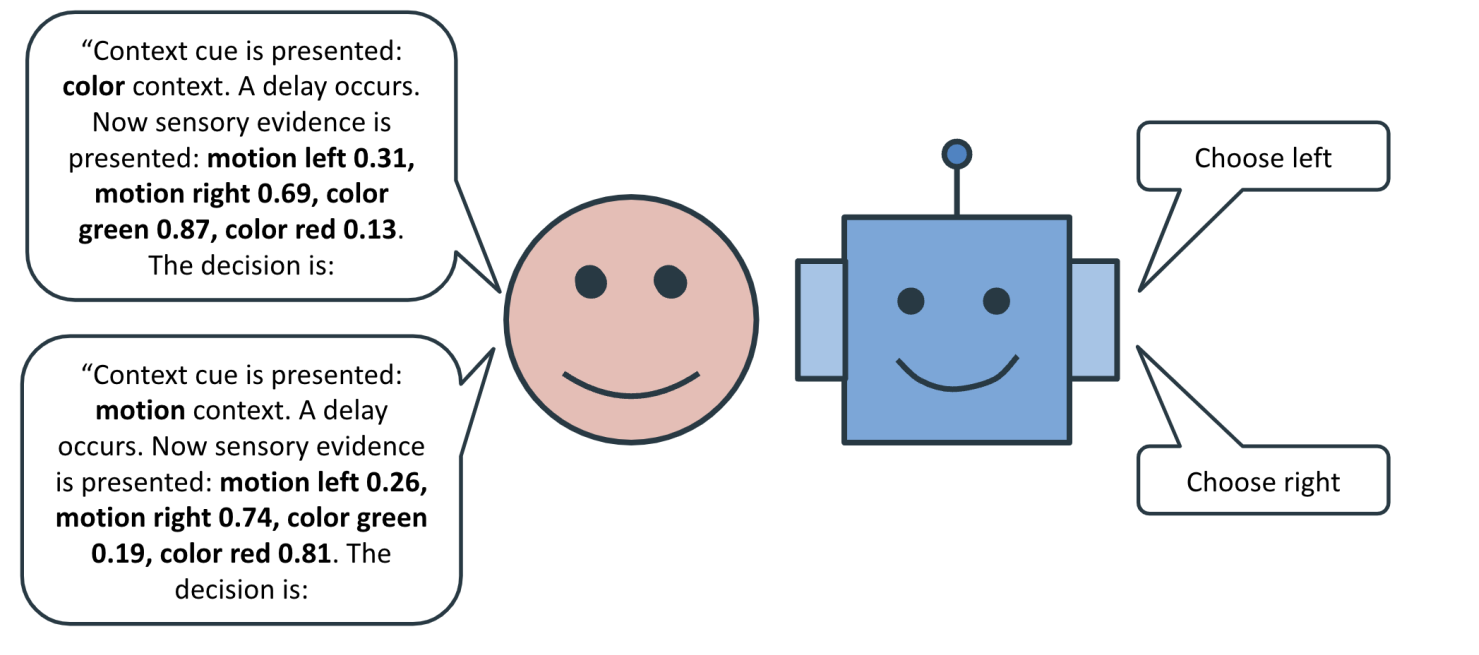}
    \caption{Task schema. Each trial of the context-dependent decision-making (CDDM) task is converted to text description and prompted to GPT-2. The model is trained to respond based on the context cue and sensory evidence, similar to the animal behavioral task. }
    \label{fig:fig1 task}
    \vspace{-10pt}
\end{figure}
\section{Related Work}
\textbf{Fine-tuning pretrained transformer models enables a variety of capabilities.} Fine-tuning pretrained transformer models has demonstrated a wide range of capabilities. For instance, LLMs such as GPT-3 can be fine-tuned to perform specific tasks like text summarization, question answering, and code generation with remarkable proficiency \citep{brown_language_2020, radford_language_2019}. This adaptability is largely attributed to the rich representations developed during the extensive pretraining phase, which captures various syntactic and semantic aspects of language \citep{devlin_bert_2019, raffel_exploring_2020}. Fine-tuning enables these models to leverage their broad knowledge base and adapt to niche tasks with relatively few task-specific examples \citep{brown_language_2020, raffel_exploring_2020}. Moreover, fine-tuned models have shown effectiveness in domains beyond natural language processing, such as protein structure prediction and mathematical theorem proving, highlighting their versatility and potential for cross-disciplinary applications \citep{jumper_highly_2021, polu_generative_2020}.

\textbf{Transformers’ relation to neuroscience.} In addition to their capability and scalability, transformers \citep{vaswani_attention_2017} have been shown to bear resemblance to traditional computational neuroscience models, namely recurrent neural networks. For example, \citet{whittington_relating_2022} showed that transformers with recurrent position encodings are similar to Hopfield-network-based models of the hippocampus, and exhibited spatial representations such as place cells and grid cells. \citet{zucchet_gated_2023} showed that linearized self-attention can be implemented in linear gated recurrent neural networks. More recent work in neurobiology even suggested that self-attention could in principle be implemented by the brain through cortical waves \citep{muller_transformers_2024} or Hebbian plasticity \citep{ellwood_short-term_2024}. As such, the idea that transformer-based LLMs can be operating on similar principles as human intelligence is not so far-fetched. These parallels also inspired our approach of analyzing transformer hidden states with standard neuroscience data analyses.

\section{Experimental Setup}
We converted the perceptual random-dot task, which is commonly used to study decision-making in higher cortical areas such as prefrontal cortex \citep{mante_context-dependent_2013}, to text description as the input to GPT-2 (Figure \ref{fig:fig1 task}A). In each trial, depending on the context cue given at the beginning of the trial,  the model is required to choose the left or right target matching the prevalent sensory evidence in either motion (if the context cue is motion) or color (if the context cue is color). As such, it requires attending to the context throughout the trial, and comparing the numerical values for the sensory evidence. Consistent with the neuroscience literature, the magnitude of the motion and color evidence in either direction is randomly controlled by two latent variables, motion coherence $\text{coh}_m$ and color coherence $\text{coh}_c$, which vary from trial to trial and are sampled from uniform distribution $U(-B, B)$. For each coherence level, the motion and color inputs are given by: 
\begin{align}
    v_{\text{motion, left}} &= \frac{1-\text{coh}_m}{2}, & v_{\text{motion, right}} &= \frac{1+\text{coh}_m}{2} \\
    v_{\text{color, green}} &= \frac{1-\text{coh}_c}{2}, & v_{\text{color, red}} &= \frac{1+\text{coh}_c}{2}
\end{align}
$B$ (“bound”) is used as a hyperparameter; given a fixed number of training samples, the smaller the bound is, the more likely the model will be trained on the same prompt more than once. Under these definitions, positive motion and color coherence provide evidence for the right choice, and negative motion and color coherence provide evidence for the left choice. This means that the model should respond “Choose left” if evidence is stronger in left (if motion context) or green (if color context), and “Choose right” if evidence is stronger in right (if motion context) or red (if color context).
 \begin{figure}[h]
    \centering
     \includegraphics[scale=0.8]{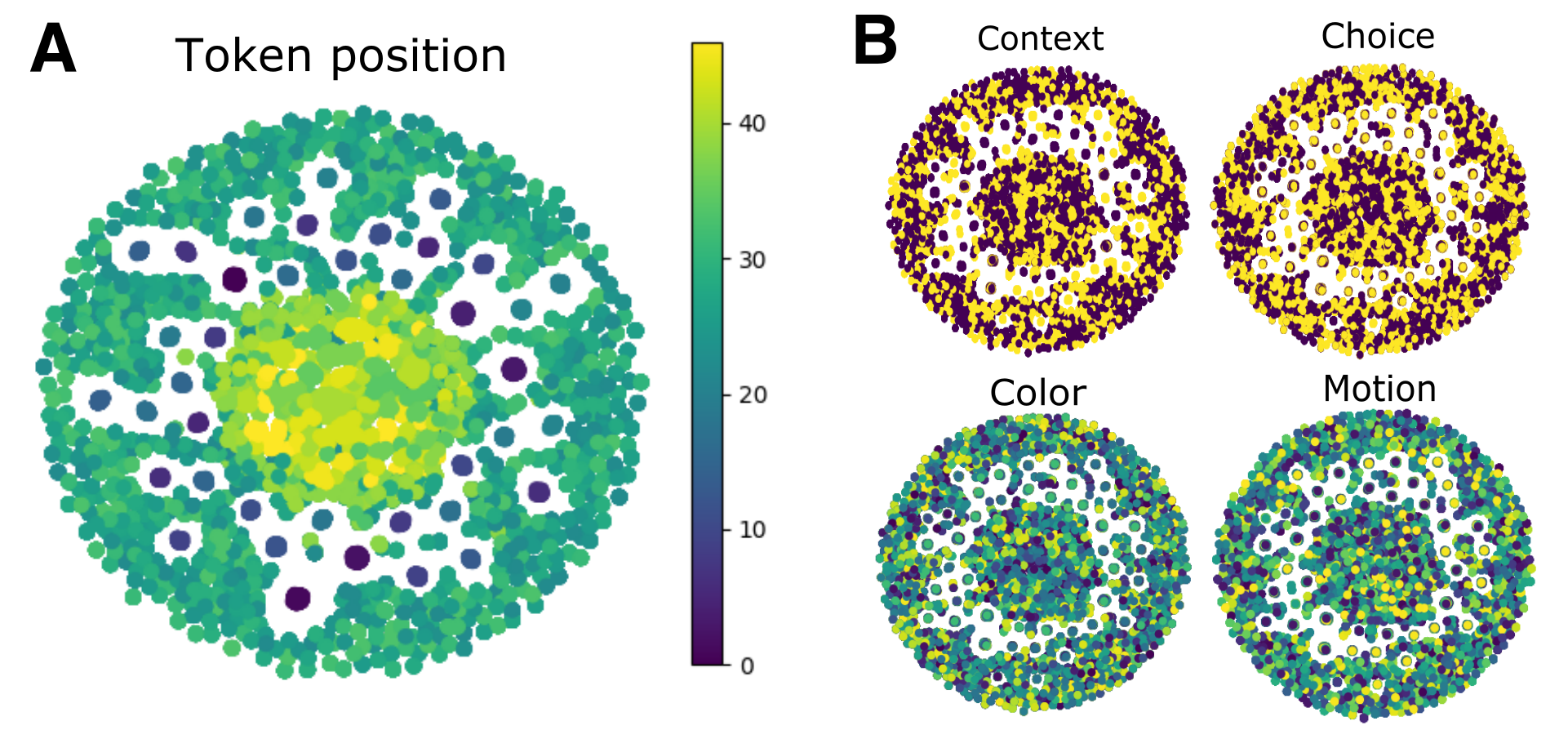}
    \caption{Last layer hidden states from different trials at different tokens after dimensionality reduction by UMAP, colored by token position (A) or different values of each behavioral variable (shown on top of each panel) (B)}
    \label{fig:fig2 UMAP}
    \vspace{-10pt}
\end{figure}
We used GPT-2 ("GPT2LMHeadModel") from the Hugging Face library \citep{radford_language_2019}, which is characterized by its 12 layers, each containing 12 attention heads, and an embedding dimension of 768, consisting of 117M parameters This model is able to produce rich behavior and representations under supervised fine-tuning, but lacks the scale and hence ability to learn this task with in-context learning.

We took GPT-2 pretrained on large corpus of English data, and fine-tuned it on this task for a range of training prompts: \{100, 500, 1000, 2000, 4000, 8000, 10000\}, and evaluated them on 2000 prompts, during which attention and hidden states data were collected. To test for generalization, we also evaluated fine-tuned models on prompts generated with different bounds (Figure \ref{fig:supp fig 1 perf}). To understand to what extent fine-tuned GPT-2 relies on pre-trained representations to solve the task, we trained GPT-2 from scratch on the same task for comparison. We found that it took 10 times more training samples to get the same level of performance from a model trained from scratch than from a fine-tuned model. We collected and analyzed attention weights ($\text{softmax}\left(\frac{QK^T}{\sqrt{d}}\right)$) and attention outputs ($\text{softmax}\left(\frac{QK^T}{\sqrt{d}}\right) \cdot V$) from each head in each layer, as well as the hidden states output from each layer. We applied various analyses to understand the population-level as well as circuit-level mechanisms by which our models solved this task (See Appendix for more details).
\section{Results}
For the fine-tuning process, we observed that the model achieves higher accuracy (measured by the percentage of correct choice) and lower loss when trained on more samples. Also, smaller bounds on the distribution from which motion and color coherence are sampled leads to better performance, suggesting that seeing repeated examples helped the model learn (Figure \ref{fig:supp fig 1 perf} top). We also found that fine-tuned models can generalize well: a model's accuracy when tested on different bounds is closely correlated with its accuracy on training data, meaning that the model learned \textit{how} to perform the task instead of simply memorizing the examples (Figure \ref{fig:supp fig 1 perf} bottom).
 \begin{figure}[h]
    \centering
     \includegraphics[scale=0.8]{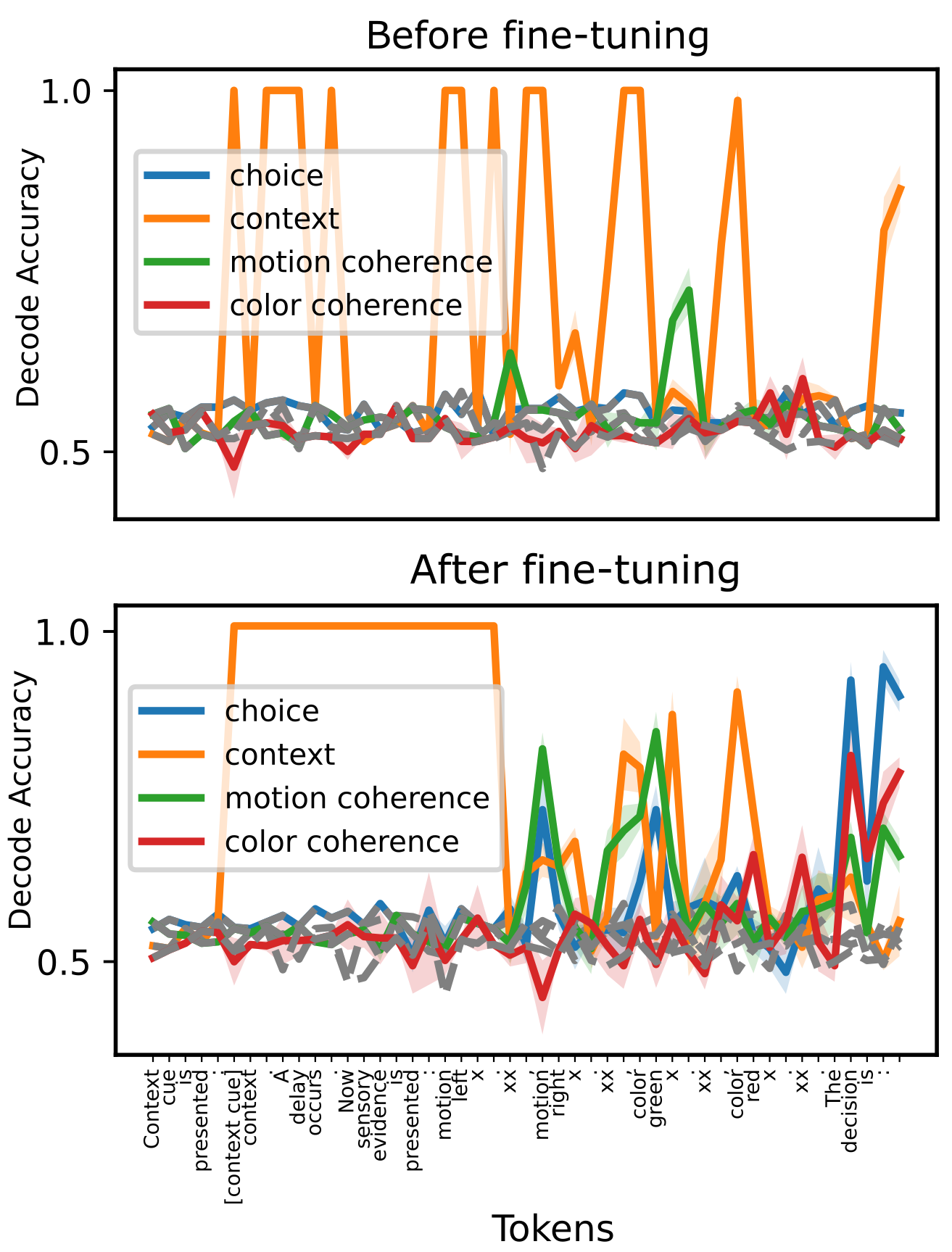}
    \caption{Accuracy of decoding each task-relevant variable from the hidden states at different tokens, for one example unit. Grey dashed lines indicate label-shuffled decoding accuracy as a baseline. Solid lines and shaded area indicate mean and standard deviation of decoding accuracy across 5 cross-validation folds. More example units shown in Figure \ref{fig:supp fig 2 mixed selectivity}}
    \label{fig:fig3 mixed selectivity}
    \vspace{-10pt}
\end{figure}
\begin{figure}[h]
    \centering
     \includegraphics[scale=0.6]{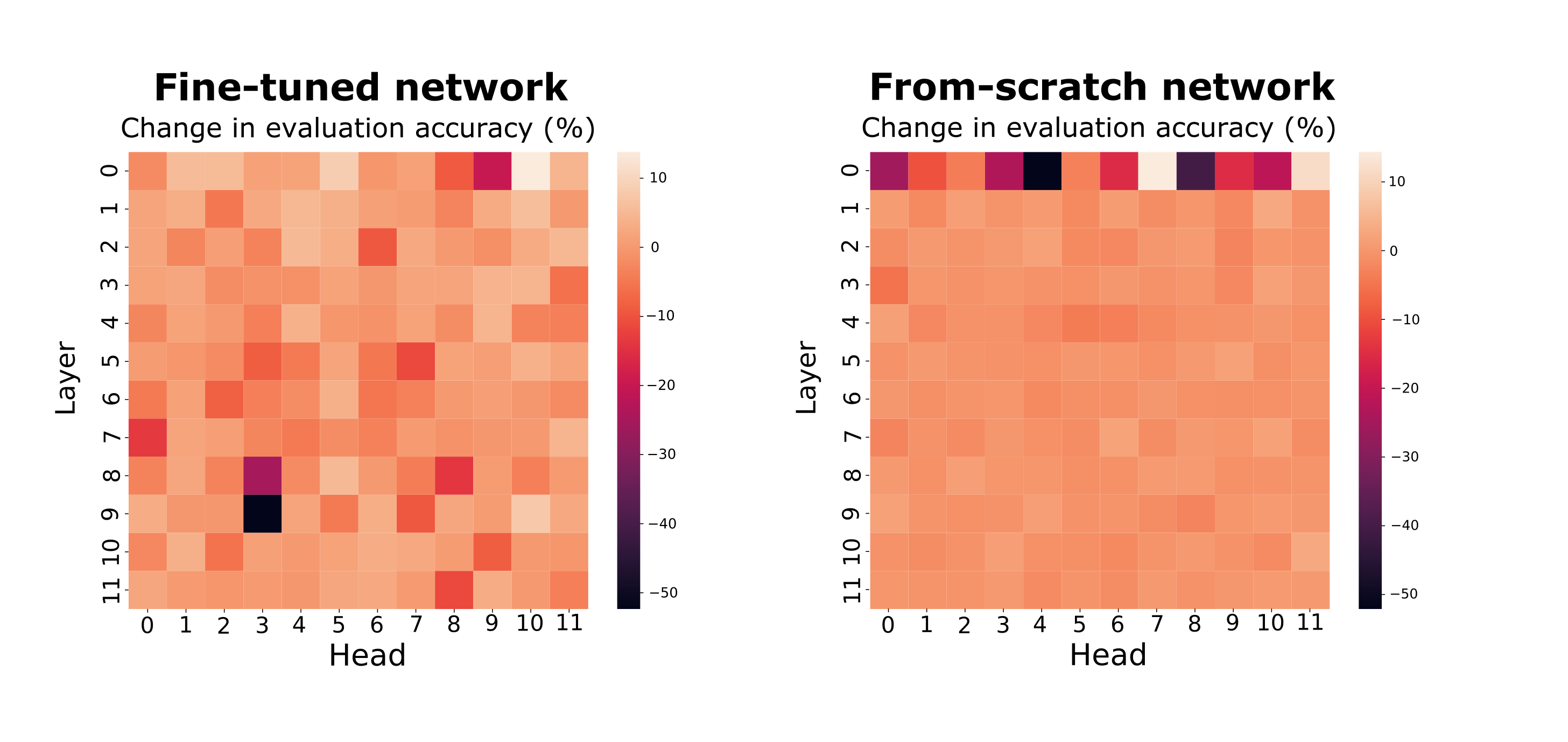}
    \caption{Evaluation accuracy, measured in the percentage of correct repsonses, after zero-ablating each attention head for a pretrained network fine-tuned on the task (left) and a network trained on the task from scratch (right).}
    \label{fig:fig4 ablation}
    \vspace{-10pt}
\end{figure}
 \begin{figure}[h]
    \centering
     \includegraphics[scale=0.4]{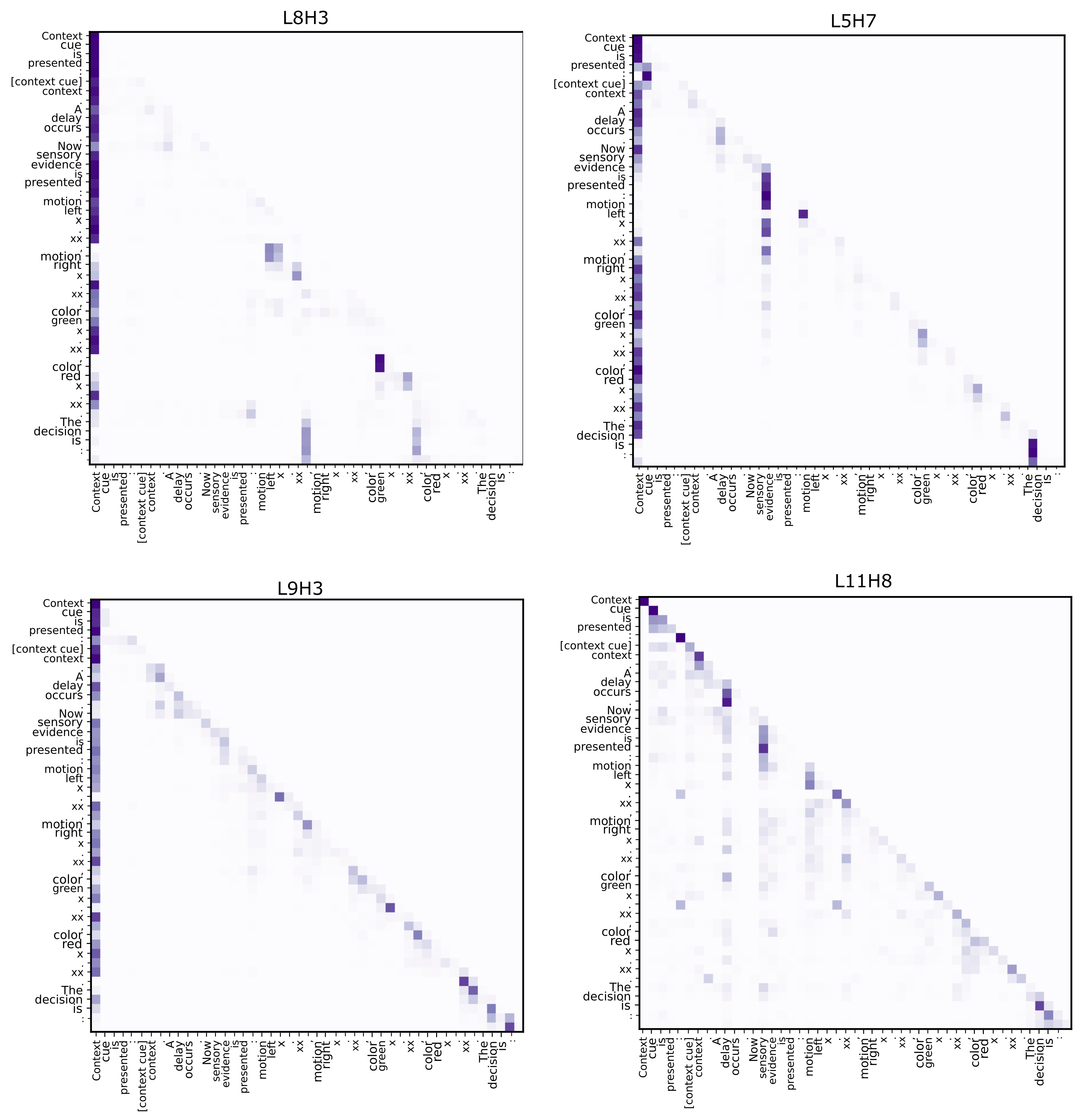}
    \caption{Average attention weights ($\text{softmax}\left(\frac{QK^T}{\sqrt{d}}\right)$) across prompts \textbf{from a fine-tuned network} for 4 example attention heads that caused significant performance drop when zero-ablated. y-axis is source token, x-axis is destination token. Darker color indicates larger value.}
    \label{fig:fig5 sft attn}
    \vspace{-10pt}
\end{figure}
\subsection{Population-level structure develops in the hidden states of fine-tuned network.} 
To understand the key driving factors of representations formed, we projected the hidden states from the last layer at all tokens in all trials to a two-dimensional space with UMAP. We found that the hidden states move in latent space according to the token position: in each prompt, the hidden states start in the middle of the UMAP space, then move outwards, then move inwards. This temporal structure does not apparently correlate to any behavioral variables (context, motion coherence, color coherence, or choice) (Figure \ref{fig:fig2 UMAP}). We decoded task-relevant variables, including context, motion coherence (whether it’s positive or negative), color coherence (whether it’s positive or negative), and choice, from the hidden states of the last layer during the presentation of each token with logistic regression before and after tine-tuning (Figure \ref{fig:fig3 mixed selectivity}; more examples shown in Figure \ref{fig:supp fig 2 mixed selectivity}). We notice that, after fine-tuning, a single unit’s activation can encode multiple independent task-relevant variables at the same time, similar to the mixed selectivity phenomenon in the brain \citep{rigotti_importance_2013}. Increasing amount of information relevant for decision-making gets encoded as token position progresses.
  \begin{figure}[h]
    \centering
     \includegraphics[scale=0.6]{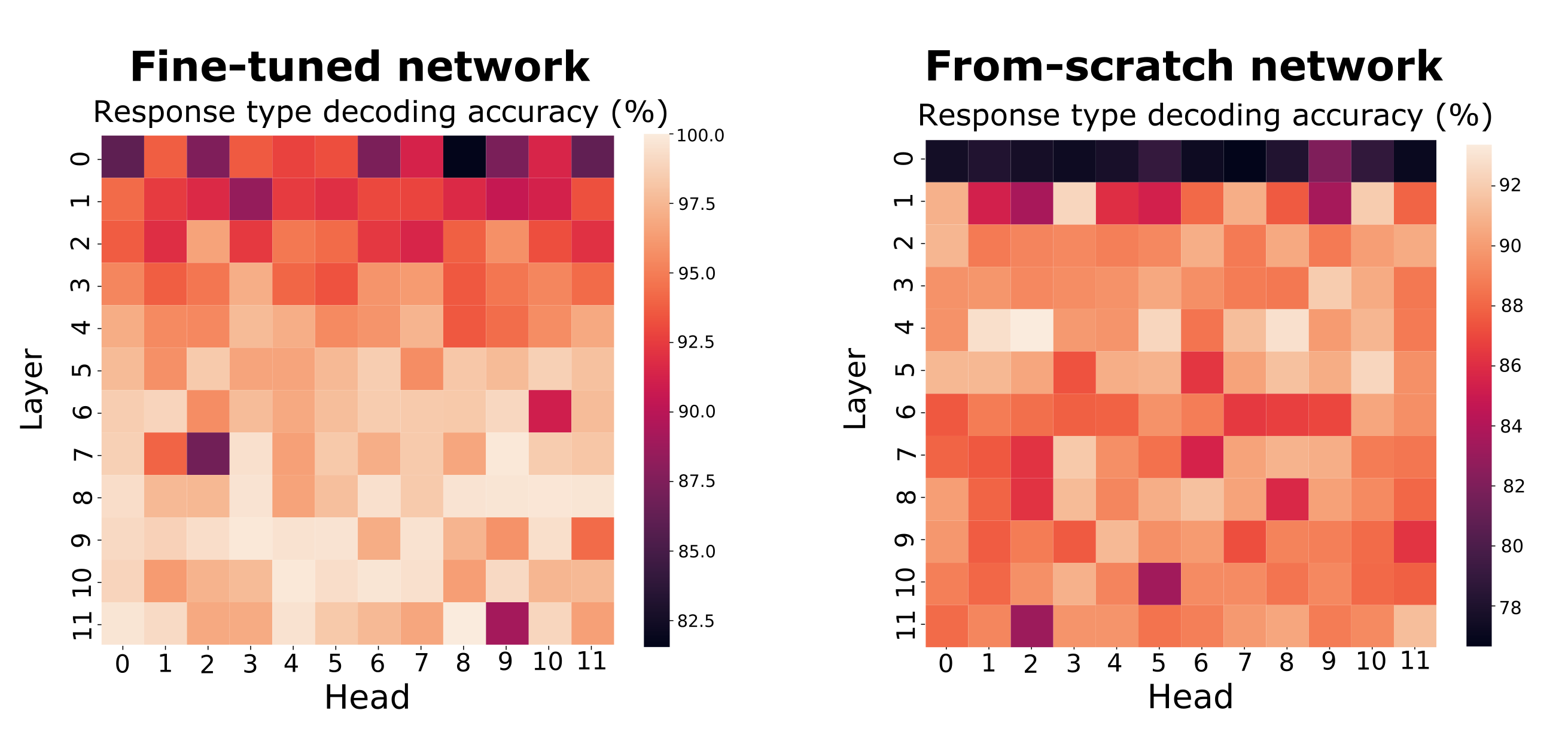}
    \caption{Accuracy of decoding the response type from the output of each attention head ($\text{softmax}\left(\frac{QK^T}{\sqrt{d}}\right) \cdot V$), in a pretrained network fine-tuned on the task (left) and a network trained on the task from scratch (right).}
    \label{fig:fig6 head output decoding}
    \vspace{-10pt}
\end{figure}
\subsection{Task-specific solutions emerge in networks trained from scratch.} 
To understand whether task-optimized networks and fine-tuned networks converge onto similar circuit-level mechanisms, we ablated each attention head in each layer in trained networks by setting their attention weight to zero, and evaluated the accuracy. We found that fine-tuned networks rely more on attention heads in later layers, particular examples included L5H7, L8H3, L9H3, L11H8, etc, where performance on the task had a significant drop after ablation (Figure \ref{fig:fig4 ablation} left). The importance of these heads is universal across models fine-tuned with different hyperparameters (Figure \ref{fig:supp fig 3 more ablation}). The trial-averaged attention patterns from these heads did not expose particularly task-specific solutions; instead, we hypothesized that these heads are important for generic language modeling and largely developed their representation during pretraining (Figure \ref{fig:fig5 sft attn}). Indeed, the same heads in models trained from scratch had not yet developed the same generic attention pattern (Figure \ref{fig:supp fig 4 train attn later layer}). In contrast, networks optimized on the task from random initialization developed more task-specific patterns in their attention weights: drastic performance change happened only when ablating heads from the first layer but not later layers (Figure \ref{fig:fig4 ablation} right). We found that first layer heads pay attention to numbers in the prompt (Figure \ref{fig:supp fig 5 train attn first layer}), which could be a task-specific solution that networks trained from scratch developed. 

Next, we asked, what about these heads makes their ablation more detrimental than other heads? To this end, we connected results from head ablation to the representation of behavior in attention outputs. For each head in each layer, we decoded with SVM the type of response produced by the model following the prompt: whether it’s choosing left, choosing right, or an invalid response (neither choosing left or right). We found that, for fine-tuned models, heads in later layers indeed showed higher decodability for response type, matching the intuition that they are more important for generic language-based tasks than earlier layers. For models trained on task from scratch, we found that the first layer, whose ablation led to significant performance drop, showed less decodability than other layers. This hints at the possibility that, while heads in the first layer pay significant attention to numbers and other relevant information, the cognitive processing and manipulation of such information to produce appropriate responses could happen in later layers (Figure \ref{fig:fig6 head output decoding}).
\section{Discussion}
This study demonstrates that fine-tuned models rely more on pretrained representations to solve a novel decision-making task, while models optimized from scratch develop alternative mechanisms. Specifically, the fine-tuned models showed significant reliance on attention heads in later layers, which are likely crucial for generic language modeling, as these heads were developed during pretraining. In contrast, models trained from scratch developed task-specific solutions, with significant performance drops upon ablating heads in the first layer, suggesting that these heads are vital for extracting task-relevant numerical information.

Our findings suggest that while fine-tuned models benefit from the pretrained representations for generalization and performance, models trained from scratch might offer more specialized and potentially efficient solutions for specific tasks. This distinction could be crucial for applications requiring highly specialized problem-solving capabilities. Future research could explore the implications of these findings for the design and deployment of LLMs in various real-world applications, considering the trade-offs between leveraging pretrained knowledge and developing task-specific solutions from scratch.

Our approach of adapting cognitive tasks commonly used in neuroscience and cognitive science laboratories also served another purpose: to evaluate the representation alignment between biological and artificial intelligence, and to advance our understanding of the general principles of intelligence. Complex intelligence systems such as the brain or its computational models often need to pack high-dimensional information from the real world into their activation, making understanding the mechanism by which these activation forms a model of the world a challenge. In the realm of AI, the phenomenon that one neuron can reliably respond to multiple semantic concepts is termed “polysemanticity”, and has been observed in vision models and the multi-layer perceptron (MLP) units in language models \citep{olah_feature_2017, olah_zoom_2020, elhage_toy_2022, scherlis_polysemanticity_2023}. Such phenomenon is similar to the neuroscience phenomenon of “mixed selectivity” in which neurons respond to a combination of sensory inputs, task conditions, and behavioral outputs. Thus, we propose that understanding how LLMs solve cognitive tasks and deciphering the representational alignment between the brain and LLMs would be crucial for the development of more robust and adaptable AI systems.
\section{Limitations and next steps}
Here we have demonstrated that fine-tuned models relied more on pretrained representations to solve a novel decision-making task, and that models optimized on the same task from scratch which do not have pretrained representations to rely on developed other mechanisms, which we hypothesize to be a combination of first layer extracting numbers and later layers manipulating these information to make a decision. However, further investigation is needed to elucidate the exact mechanism by which models trained from scratch solve this task. Another limitation is that we drew our conclusions based on one cognitive task; further studies with more diverse cognitive tasks are required to understand how pretrained representations support task-specific fine-tuning. Lastly, it is important to develop new quantitative metrics to ensure scientific rigor in our results, as much of our current findings are based on qualitative observations. Similarly, the field of mechanistic interpretability in LLMs, which is also largely qualitative at present, requires new quantitative methods to advance.




\section*{Acknowledgements}

D.L. is supported by a NSERC Canada Graduate Scholarship—Doctoral (569390-2022). This research was enabled in part by support provided by Calcul Québec (https://www.calculquebec.ca/en/) and Compute Canada (www.computecanada.ca). The author acknowledges the material support of NVIDIA in the form of computational resources.

\bibliography{ICML2024LLMCog}

\begin{thebibliography}{17}
\providecommand{\natexlab}[1]{#1}
\providecommand{\url}[1]{\texttt{#1}}
\expandafter\ifx\csname urlstyle\endcsname\relax
  \providecommand{\doi}[1]{doi: #1}\else
  \providecommand{\doi}{doi: \begingroup \urlstyle{rm}\Url}\fi

\bibitem[Brown et~al.(2020)Brown, Mann, Ryder, Subbiah, Kaplan, Dhariwal, Neelakantan, Shyam, Sastry, Askell, Agarwal, Herbert-Voss, Krueger, Henighan, Child, Ramesh, Ziegler, Wu, Winter, Hesse, Chen, Sigler, Litwin, Gray, Chess, Clark, Berner, McCandlish, Radford, Sutskever, and Amodei]{brown_language_2020}
Brown, T., Mann, B., Ryder, N., Subbiah, M., Kaplan, J.~D., Dhariwal, P., Neelakantan, A., Shyam, P., Sastry, G., Askell, A., Agarwal, S., Herbert-Voss, A., Krueger, G., Henighan, T., Child, R., Ramesh, A., Ziegler, D., Wu, J., Winter, C., Hesse, C., Chen, M., Sigler, E., Litwin, M., Gray, S., Chess, B., Clark, J., Berner, C., McCandlish, S., Radford, A., Sutskever, I., and Amodei, D.
\newblock Language {Models} are {Few}-{Shot} {Learners}.
\newblock In \emph{Advances in {Neural} {Information} {Processing} {Systems}}, volume~33, pp.\  1877--1901. Curran Associates, Inc., 2020.
\newblock URL \url{https://papers.nips.cc/paper_files/paper/2020/hash/1457c0d6bfcb4967418bfb8ac142f64a-Abstract.html}.

\bibitem[Devlin et~al.(2019)Devlin, Chang, Lee, and Toutanova]{devlin_bert_2019}
Devlin, J., Chang, M.-W., Lee, K., and Toutanova, K.
\newblock {BERT}: {Pre}-training of {Deep} {Bidirectional} {Transformers} for {Language} {Understanding}, May 2019.
\newblock URL \url{http://arxiv.org/abs/1810.04805}.
\newblock arXiv:1810.04805 [cs].

\bibitem[Elhage et~al.(2022)Elhage, Hume, Olsson, Schiefer, Henighan, Kravec, Hatfield-Dodds, Lasenby, Drain, Chen, Grosse, McCandlish, Kaplan, Amodei, Wattenberg, and Olah]{elhage_toy_2022}
Elhage, N., Hume, T., Olsson, C., Schiefer, N., Henighan, T., Kravec, S., Hatfield-Dodds, Z., Lasenby, R., Drain, D., Chen, C., Grosse, R., McCandlish, S., Kaplan, J., Amodei, D., Wattenberg, M., and Olah, C.
\newblock Toy {Models} of {Superposition}, September 2022.
\newblock URL \url{http://arxiv.org/abs/2209.10652}.
\newblock arXiv:2209.10652 [cs].

\bibitem[Ellwood(2024)]{ellwood_short-term_2024}
Ellwood, I.~T.
\newblock Short-term {Hebbian} learning can implement transformer-like attention.
\newblock \emph{PLOS Computational Biology}, 20\penalty0 (1):\penalty0 e1011843, January 2024.
\newblock ISSN 1553-7358.
\newblock \doi{10.1371/journal.pcbi.1011843}.
\newblock URL \url{https://journals.plos.org/ploscompbiol/article?id=10.1371/journal.pcbi.1011843}.
\newblock Publisher: Public Library of Science.

\bibitem[Jumper et~al.(2021)Jumper, Evans, Pritzel, Green, Figurnov, Ronneberger, Tunyasuvunakool, Bates, Žídek, Potapenko, Bridgland, Meyer, Kohl, Ballard, Cowie, Romera-Paredes, Nikolov, Jain, Adler, Back, Petersen, Reiman, Clancy, Zielinski, Steinegger, Pacholska, Berghammer, Bodenstein, Silver, Vinyals, Senior, Kavukcuoglu, Kohli, and Hassabis]{jumper_highly_2021}
Jumper, J., Evans, R., Pritzel, A., Green, T., Figurnov, M., Ronneberger, O., Tunyasuvunakool, K., Bates, R., Žídek, A., Potapenko, A., Bridgland, A., Meyer, C., Kohl, S. A.~A., Ballard, A.~J., Cowie, A., Romera-Paredes, B., Nikolov, S., Jain, R., Adler, J., Back, T., Petersen, S., Reiman, D., Clancy, E., Zielinski, M., Steinegger, M., Pacholska, M., Berghammer, T., Bodenstein, S., Silver, D., Vinyals, O., Senior, A.~W., Kavukcuoglu, K., Kohli, P., and Hassabis, D.
\newblock Highly accurate protein structure prediction with {AlphaFold}.
\newblock \emph{Nature}, 596\penalty0 (7873):\penalty0 583--589, August 2021.
\newblock ISSN 1476-4687.
\newblock \doi{10.1038/s41586-021-03819-2}.
\newblock URL \url{https://www.nature.com/articles/s41586-021-03819-2}.
\newblock Publisher: Nature Publishing Group.

\bibitem[Mante et~al.(2013)Mante, Sussillo, Shenoy, and Newsome]{mante_context-dependent_2013}
Mante, V., Sussillo, D., Shenoy, K.~V., and Newsome, W.~T.
\newblock Context-dependent computation by recurrent dynamics in prefrontal cortex.
\newblock \emph{Nature}, 503\penalty0 (7474):\penalty0 78--84, November 2013.
\newblock ISSN 1476-4687.
\newblock \doi{10.1038/nature12742}.
\newblock URL \url{https://www.nature.com/articles/nature12742}.
\newblock Number: 7474 Publisher: Nature Publishing Group.

\bibitem[Muller et~al.(2024)Muller, Churchland, and Sejnowski]{muller_transformers_2024}
Muller, L., Churchland, P.~S., and Sejnowski, T.~J.
\newblock Transformers and {Cortical} {Waves}: {Encoders} for {Pulling} {In} {Context} {Across} {Time}, January 2024.
\newblock URL \url{http://arxiv.org/abs/2401.14267}.
\newblock arXiv:2401.14267 [cs].

\bibitem[Olah et~al.(2017)Olah, Mordvintsev, and Schubert]{olah_feature_2017}
Olah, C., Mordvintsev, A., and Schubert, L.
\newblock Feature {Visualization}.
\newblock \emph{Distill}, 2\penalty0 (11):\penalty0 e7, November 2017.
\newblock ISSN 2476-0757.
\newblock \doi{10.23915/distill.00007}.
\newblock URL \url{https://distill.pub/2017/feature-visualization}.

\bibitem[Olah et~al.(2020)Olah, Cammarata, Schubert, Goh, Petrov, and Carter]{olah_zoom_2020}
Olah, C., Cammarata, N., Schubert, L., Goh, G., Petrov, M., and Carter, S.
\newblock Zoom {In}: {An} {Introduction} to {Circuits}.
\newblock \emph{Distill}, 5\penalty0 (3):\penalty0 e00024.001, March 2020.
\newblock ISSN 2476-0757.
\newblock \doi{10.23915/distill.00024.001}.
\newblock URL \url{https://distill.pub/2020/circuits/zoom-in}.

\bibitem[Polu \& Sutskever(2020)Polu and Sutskever]{polu_generative_2020}
Polu, S. and Sutskever, I.
\newblock Generative {Language} {Modeling} for {Automated} {Theorem} {Proving}, September 2020.
\newblock URL \url{http://arxiv.org/abs/2009.03393}.
\newblock arXiv:2009.03393 [cs, stat].

\bibitem[Radford et~al.(2019)Radford, Wu, Child, Luan, Amodei, and Sutskever]{radford_language_2019}
Radford, A., Wu, J., Child, R., Luan, D., Amodei, D., and Sutskever, I.
\newblock Language {Models} are {Unsupervised} {Multitask} {Learners}.
\newblock 2019.

\bibitem[Raffel et~al.(2020)Raffel, Shazeer, Roberts, Lee, Narang, Matena, Zhou, Li, and Liu]{raffel_exploring_2020}
Raffel, C., Shazeer, N., Roberts, A., Lee, K., Narang, S., Matena, M., Zhou, Y., Li, W., and Liu, P.~J.
\newblock Exploring the {Limits} of {Transfer} {Learning} with a {Unified} {Text}-to-{Text} {Transformer}.
\newblock \emph{Journal of Machine Learning Research}, 21\penalty0 (140):\penalty0 1--67, 2020.
\newblock ISSN 1533-7928.
\newblock URL \url{http://jmlr.org/papers/v21/20-074.html}.

\bibitem[Rigotti et~al.(2013)Rigotti, Barak, Warden, Wang, Daw, Miller, and Fusi]{rigotti_importance_2013}
Rigotti, M., Barak, O., Warden, M.~R., Wang, X.-J., Daw, N.~D., Miller, E.~K., and Fusi, S.
\newblock The importance of mixed selectivity in complex cognitive tasks.
\newblock \emph{Nature}, 497\penalty0 (7451):\penalty0 585--590, May 2013.
\newblock ISSN 1476-4687.
\newblock \doi{10.1038/nature12160}.
\newblock URL \url{https://www.nature.com/articles/nature12160}.
\newblock Number: 7451 Publisher: Nature Publishing Group.

\bibitem[Scherlis et~al.(2023)Scherlis, Sachan, Jermyn, Benton, and Shlegeris]{scherlis_polysemanticity_2023}
Scherlis, A., Sachan, K., Jermyn, A.~S., Benton, J., and Shlegeris, B.
\newblock Polysemanticity and {Capacity} in {Neural} {Networks}, July 2023.
\newblock URL \url{http://arxiv.org/abs/2210.01892}.
\newblock arXiv:2210.01892 [cs].

\bibitem[Vaswani et~al.(2017)Vaswani, Shazeer, Parmar, Uszkoreit, Jones, Gomez, Kaiser, and Polosukhin]{vaswani_attention_2017}
Vaswani, A., Shazeer, N., Parmar, N., Uszkoreit, J., Jones, L., Gomez, A.~N., Kaiser, L., and Polosukhin, I.
\newblock Attention {Is} {All} {You} {Need}, June 2017.
\newblock URL \url{https://arxiv.org/abs/1706.03762v7}.

\bibitem[Whittington et~al.(2022)Whittington, Warren, and Behrens]{whittington_relating_2022}
Whittington, J. C.~R., Warren, J., and Behrens, T. E.~J.
\newblock Relating transformers to models and neural representations of the hippocampal formation, March 2022.
\newblock URL \url{http://arxiv.org/abs/2112.04035}.
\newblock arXiv:2112.04035 [cs, q-bio].

\bibitem[Zucchet et~al.(2023)Zucchet, Kobayashi, Akram, von Oswald, Larcher, Steger, and Sacramento]{zucchet_gated_2023}
Zucchet, N., Kobayashi, S., Akram, Y., von Oswald, J., Larcher, M., Steger, A., and Sacramento, J.
\newblock Gated recurrent neural networks discover attention, September 2023.
\newblock URL \url{http://arxiv.org/abs/2309.01775}.
\newblock arXiv:2309.01775 [cs].

\end{thebibliography}
\bibliographystyle{icml2024}

\newpage
\appendix
\onecolumn
\section{Appendix}
\subsection{Training and fine-tuning details}

To fine-tune the GPT2LMHeadModel on the CDDM task, we generated a text-based CDDM task dataset, in which each sample follows the template “"Context cue is presented: ... context. A delay occurs. Now sensory evidence is presented: motion left …, motion right …, color green …, color red…. The decision is: choose …”, and we concatenated the samples to become the train dataset for the next-token prediction task for the model. We loaded the pretrained checkpoint and tokenizer from huggingface.

To evaluate the model accuracy, we generated new samples either with the same bound for motion and color coherence, or a different bound to test generalization. We asked the model to generate the choice (“Choose left/right”) given the prompt that contains the context cue and the numerical evidence. The evaluation accuracy is defined as the percentage of responses answered correctly.

To train the model on the task from scratch, we used the same training and testing data, as well as a pretrained GPT2 tokenizer, but used newly initialized GPT2LMHeadModel without loading any weights.

For both training and fine-tuning, we ran the experiments with 3 seeds, and grid-searched through the hyperparameter space for training epochs, per-device train batch size, and learning rate. The major results shown in the paper come from the hyperparameter settings in Table \ref{tab:comparison}.

\begin{table}[h]
\centering
\begin{tabular}{| l | l | l |}
\hline
 & Fine-tuned model & From-scratch model \\
\hline
Training epochs & 12 & 50 \\
\hline
Number of training samples & 8000 & 200000 \\
\hline
Per-device train batch size & 4 & 16 \\
\hline
Learning rate & 5e-5 & 0.0001 \\
\hline
Bound for coherence & 0.9 & 0.7 \\
\hline
Seed & 2024 & 2026 \\
\hline
\end{tabular}
\caption{Hyperparameters used for fine-tuned and from-scratch models}
\label{tab:comparison}
\end{table}

\subsection{Analysis}

\subsubsection{Analysis on last layer hidden states}
During the evaluation phase when the model is asked to generate the choice based on the prompt, we collected the hidden states output (dim=768) from the transformer module at each token (in total 47 tokens in the prompt), before they were used to compute the logits.

To project the 768-dimensional vector into a two-dimensional space with UMAP, we stacked the vectors across evaluation prompts across tokens. We color coded the results from dimensionality reduction with token position, context (color or motion), choice (left or right), motion coherence (a number uniformly sampled between -bound and bound), and color coherence (a number uniformly sampled between -bound and bound).

Logistic regression decoding on behavioral variables: we trained a logistic regressor to map the 768-dimensional vector at different token positions in each prompt to the corresponding behavioral variable (context, choice, motion coherence, color coherence) of that trial. We used 5-fold cross-validation, in which in each fold, 80\% of the data is used for training and 20\% for calculating the mapping accuracy, and averaged across the folds. To obtain a baseline, we shuffled the value of behavioral variables and ran the same analysis.

\subsubsection{Analysis on attention weights and attention outputs}
Zero ablation: To assess the importance of each attention head to the task performance, during evaluation, we set the attention weights of target head to 0 during forward pass, and evaluate the accuracy on the decision-making task.

During evaluation, at each token position, we collected the attention output from each head in each layer (dim=64) before they were concatenated and used to compute the hidden states output from that attention layer. We trained SVM decoder to decode the response type (choose left, choose right, or invalid response that is either choose left or choose right) from attention outputs across tokens during each prompt (i.e., we concatenated the 64-dimensional vectors from all token positions in a prompt, and used that to decode the type of response to that prompt).

Our analysis was done using scikit-learn (for UMAP, SVM decoding, logistic regression), and we used matplotlib and seaborn libraries to generate our plots. Experiments were done using pytorch and numpy packages. We aim to maximize the reproducibility of this study, and we will release the codebase upon acceptance.

\newpage
\renewcommand{\thefigure}{A.\arabic{figure}}
\renewcommand{\theHfigure}{Appendix.\thefigure} 
\setcounter{figure}{0} 
\begin{figure}[ht]
    \centering
    \includegraphics[width=1\linewidth]{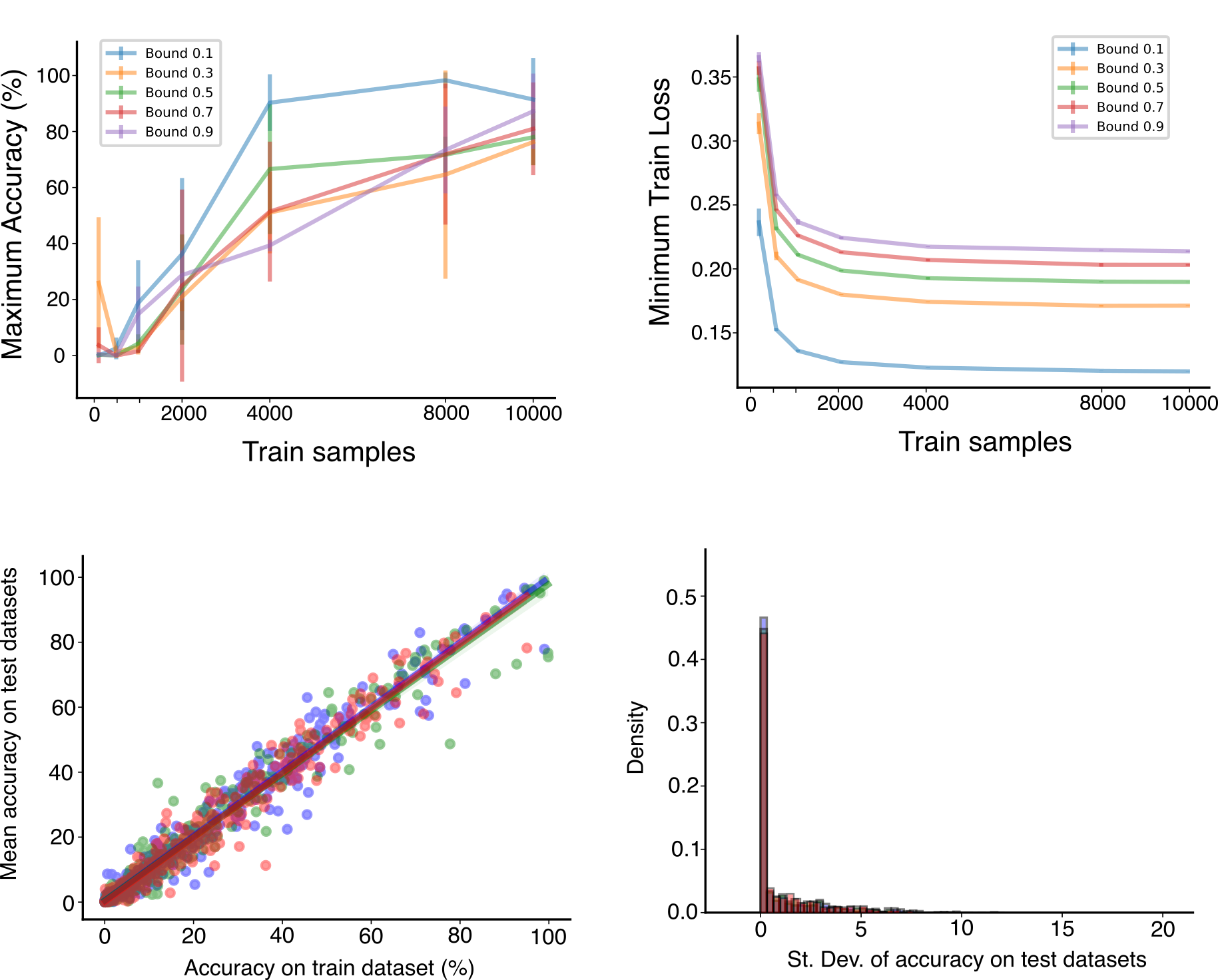}
    \caption{\textbf{Top: Performance of network after fine-tuning.} We plotted the maximum accuracy and minimum loss for each hyperparameter setting, given the number of trials used for training. Error bar shows mean and standard deviation across 3 seeds.\textbf{ Bottom: Generalization to test datasets of different bounds.} \textbf{Left:} Mean accuracy on test datasets across the 4 different test bounds that are different from the bound that the model was fine-tuned on, plotted against model’s accuracy on the train datasets. Different colors indicate different seeds for fine-tuning. \textbf{Right:} Histogram of the standard deviation of accuracy on test datasets across the 4 different test bounds.}
    \label{fig:supp fig 1 perf}
\end{figure}

\newpage
\begin{figure}[ht]
    \centering
    \includegraphics[scale=0.8]{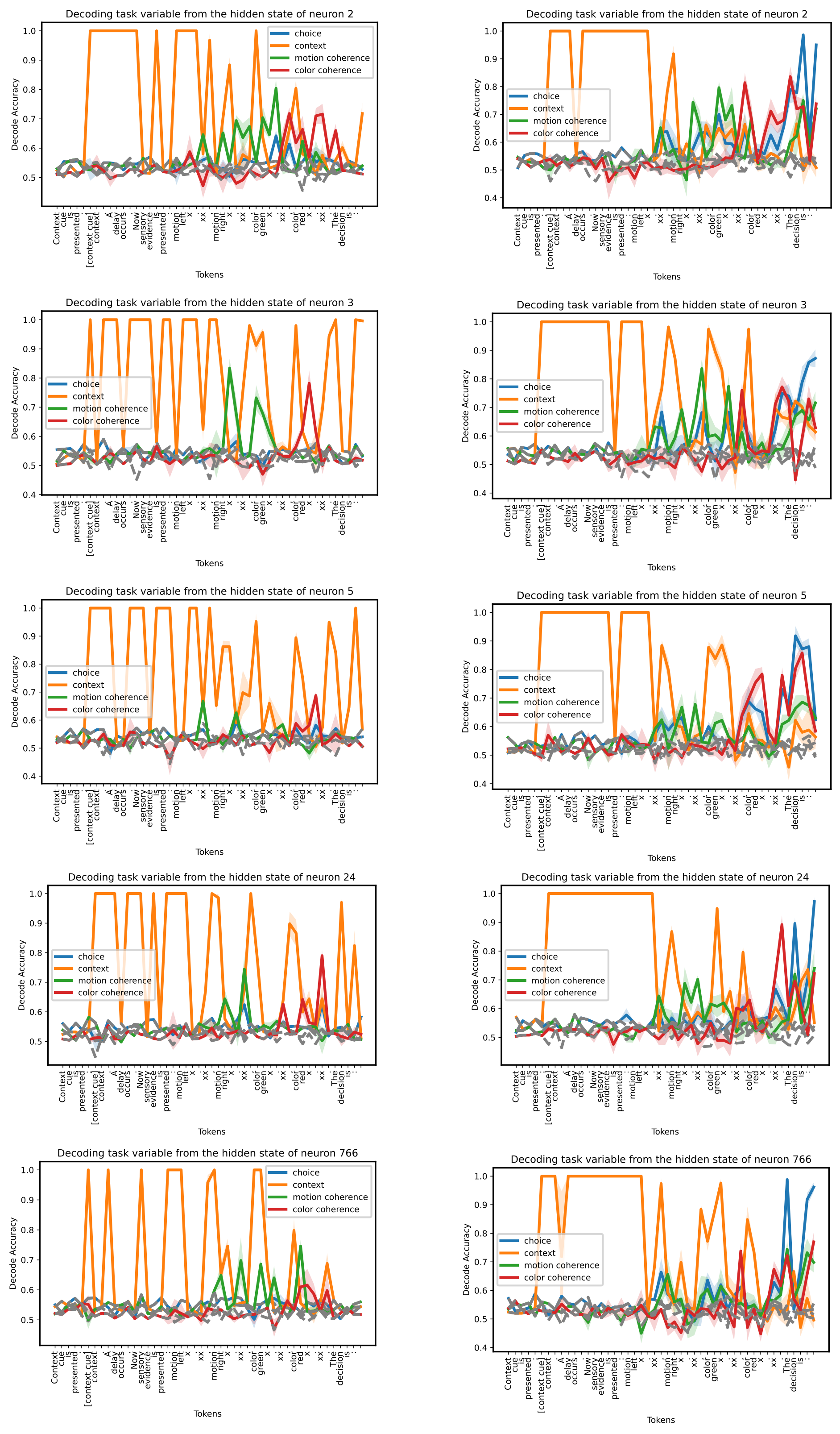}
    \caption{Behavioral variable decoding (with logistic regression) from last-layer hidden states at each token, for 5 different units (corresponding to the 5 rows). Left column: before fine-tuning. Right column: after fine-tuning.}
    \label{fig:supp fig 2 mixed selectivity}
\end{figure}

\newpage
\begin{figure}[ht]
    \centering
    \includegraphics[scale=0.8]{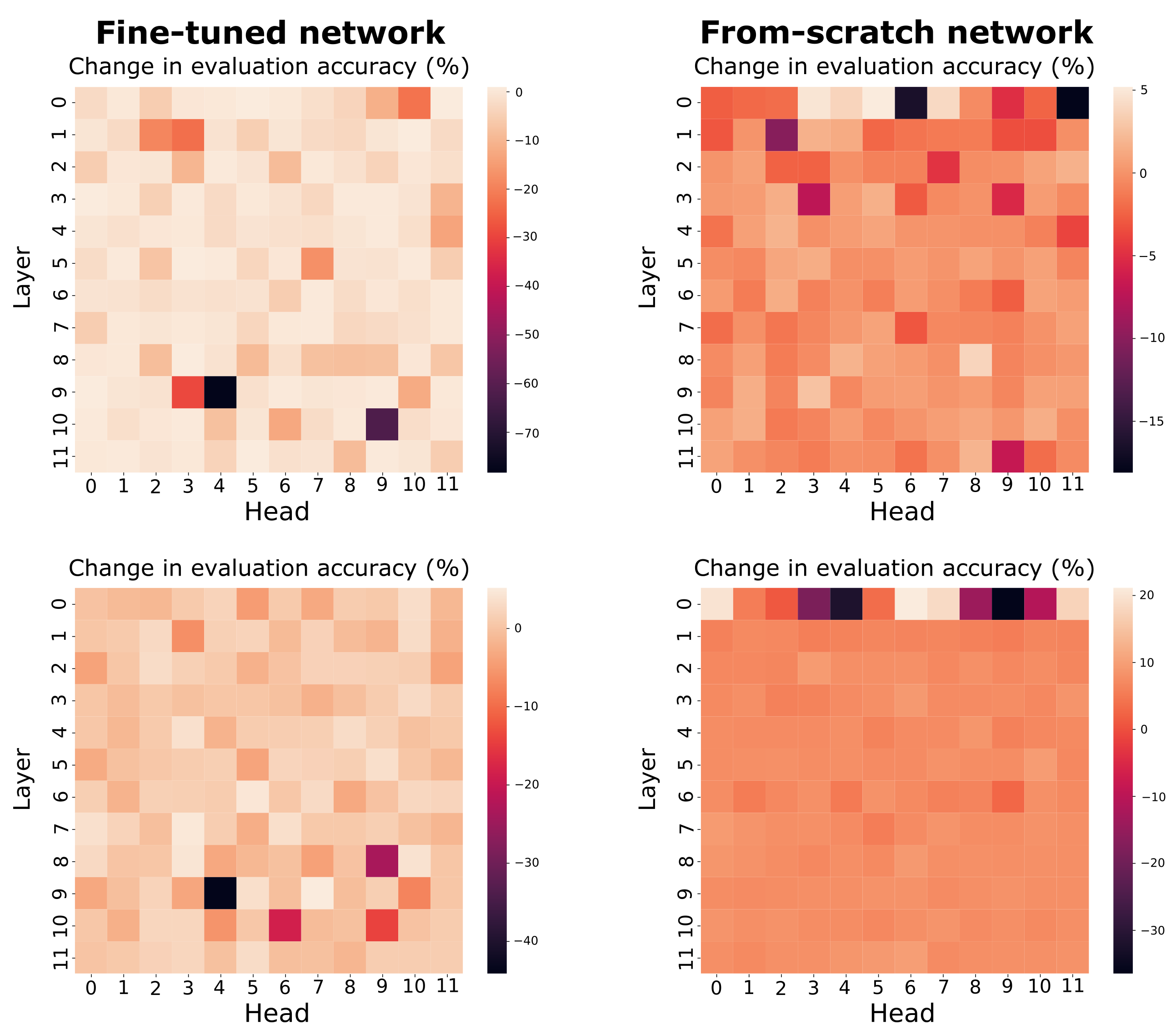}
    \caption{Supplementary ablation results: Evaluation accuracy, measured in the percentage of correct repsonses, after zero-ablating each attention head, for two more pretrained networks fine-tuned on the task (left) and two more networks trained on the task from scratch (right).}
    \label{fig:supp fig 3 more ablation}
\end{figure}

\begin{figure}[ht]
    \centering
    \includegraphics[scale=0.8]{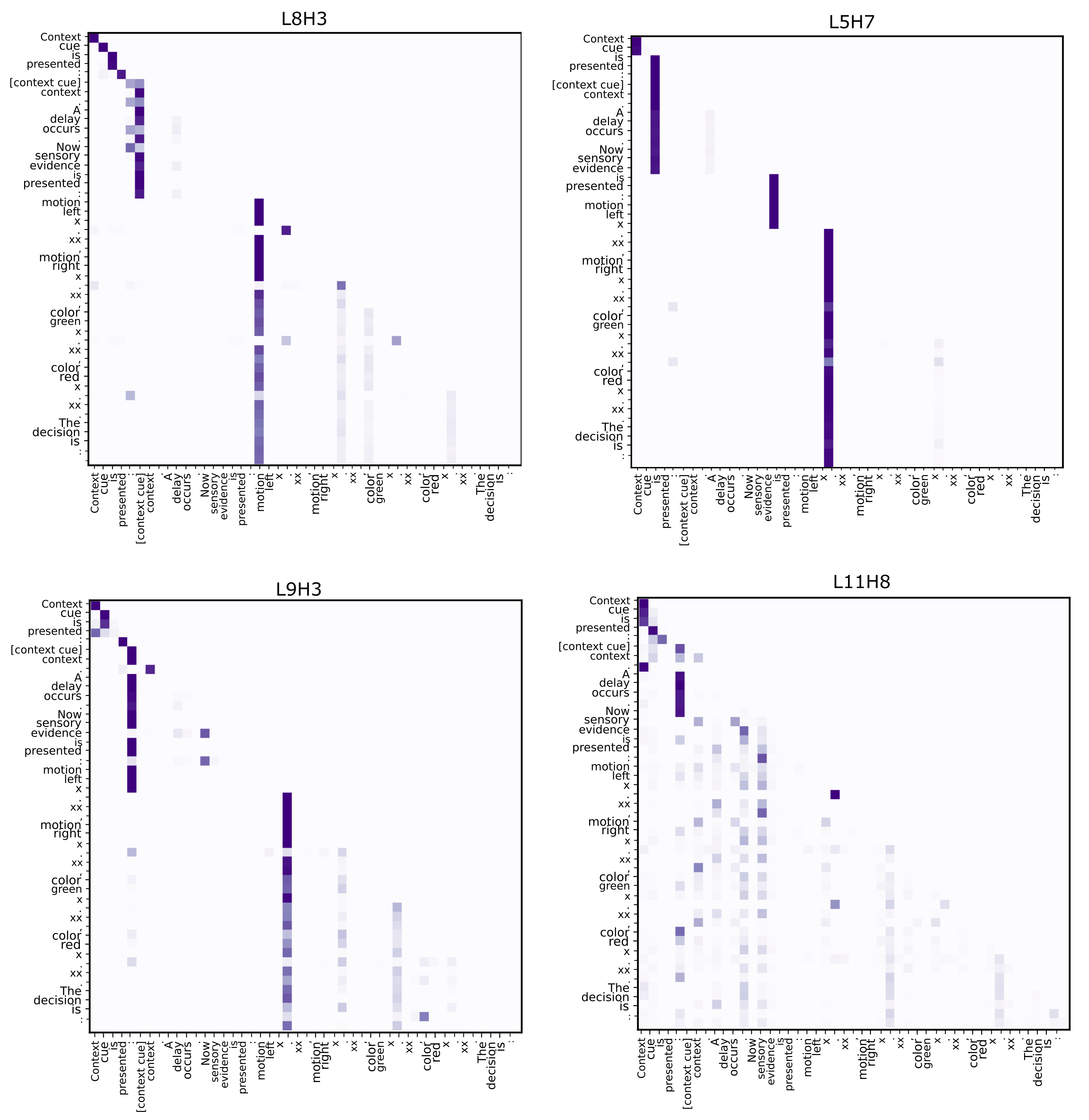}
    \caption{Average attention weights ($\text{softmax}\left(\frac{QK^T}{\sqrt{d}}\right)$) across prompts \textbf{from a network trained from scratch} for the 4 example attention heads whose zero-ablation in fine-tuned networks caused significant performance drop. y-axis is source token, x-axis is destination token. Darker color indicates larger value.}
    \label{fig:supp fig 4 train attn later layer}
\end{figure}

\newpage
\begin{figure}[ht]
    \centering
    \includegraphics[scale=0.8]{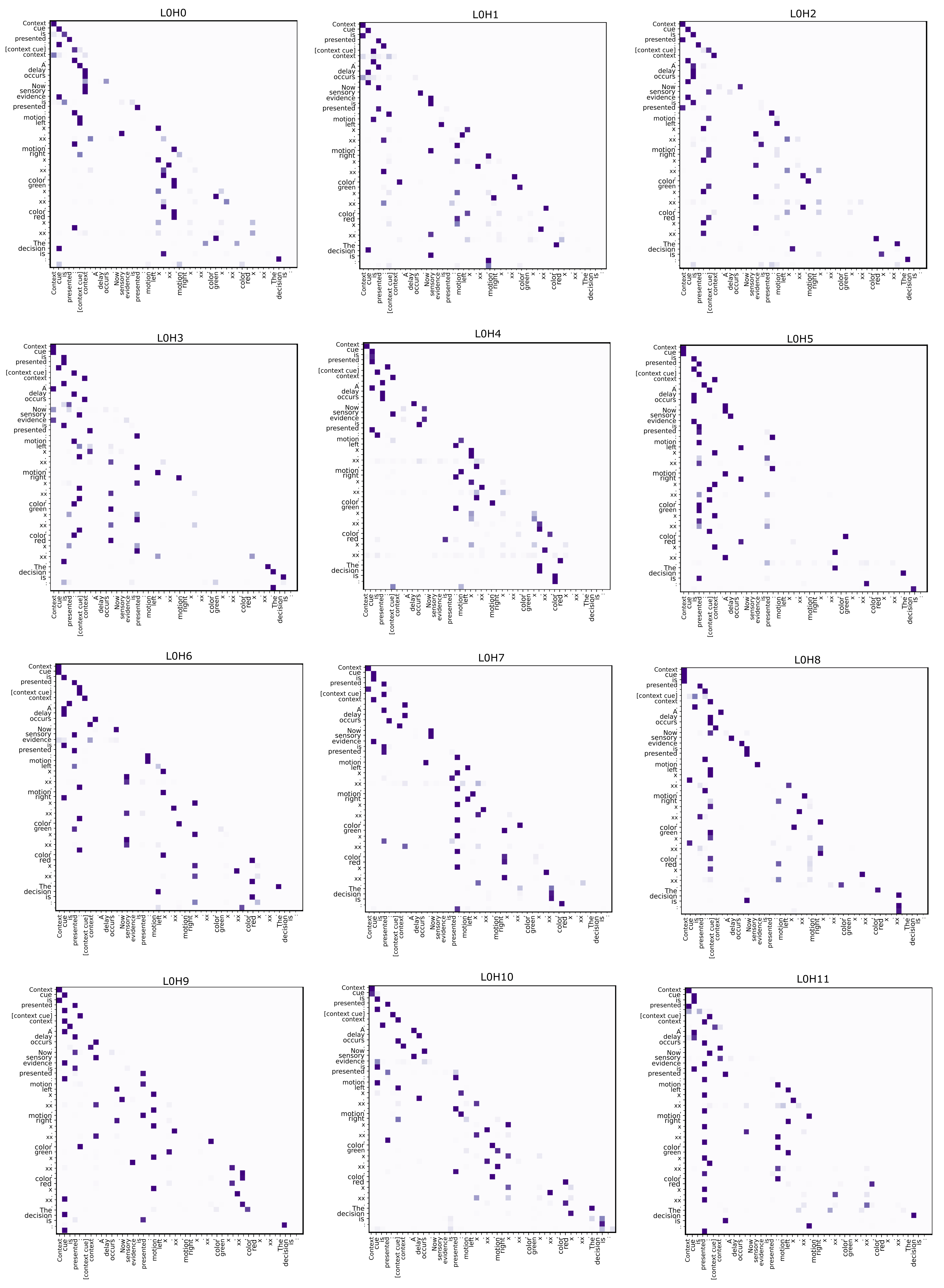}
    \caption{Average attention weights ($\text{softmax}\left(\frac{QK^T}{\sqrt{d}}\right)$) across prompts from a network trained from scratch for the 12 heads in the first layer. y-axis is source token, x-axis is destination token. Darker color indicates larger value.}
    \label{fig:supp fig 5 train attn first layer}
\end{figure}



\end{document}